\pdfoutput=1
\documentclass[11pt]{article}
\usepackage{enumitem}
\usepackage{acl}
\usepackage{times}
\usepackage{latexsym}
\usepackage{multicol}
\usepackage{graphicx} 
\usepackage{natbib}  
\usepackage{caption} 
\usepackage{amssymb}
\usepackage[ruled,linesnumbered]{algorithm2e}
\usepackage{color}
\usepackage{amsmath}
\usepackage{amsthm}
\usepackage{color}
\usepackage{adjustbox}
\usepackage{multirow}
\usepackage{booktabs}
\usepackage{amsmath}
\usepackage{hyperref}
\usepackage[inkscapelatex=false]{svg}
\usepackage{subfigure}
\newcommand*{\rom}[1]{\expandafter\@slowromancap\romannumeral #1@}

\usepackage[T1]{fontenc}

\usepackage[utf8]{inputenc}

\usepackage{microtype}
\title{On the Evaluation Metrics for Paraphrase Generation}
\author{Lingfeng Shen \\ Johns Hopkins University
        \And Lemao Liu\\Tencent AI Lab \And Haiyun Jiang\\Tencent AI Lab\\ \And
        Shuming Shi \\ Tenent AI Lab}

\begin{document}
\maketitle
\begin{abstract}

In this paper we revisit automatic metrics for paraphrase evaluation and obtain two findings that disobey conventional wisdom: (1) Reference-free metrics achieve better performance than their reference-based counterparts. (2) Most commonly used metrics do not align well with human annotation.
Underlying reasons behind the above findings are explored through additional experiments and in-depth analyses.
Based on the experiments and analyses, we propose ParaScore, a new evaluation metric for paraphrase generation. It possesses the merits of reference-based and reference-free metrics and explicitly models lexical divergence.
Experimental results demonstrate that ParaScore significantly outperforms existing metrics. The codes and toolkit are released in \url{https://github.com/shadowkiller33/ParaScore}.
\end{abstract}

\section{Introduction}
Paraphrase generation is a fundamental problem in natural language processing (NLP), which has been widely applied in versatile tasks, such as question answering \cite{dong2017learning,lan2018neural,gan2019improving,abujabal2019comqa}, machine translation \cite{madnani2012re,apidianaki2018automated,kajiwara2019negative}, and semantic parsing \cite{herzig2019don,wu2021paraphrasing,cao2020unsupervised}.
Recent years have witnessed rapid development in paraphrase generation algorithms~\cite{sun2021aesop,huang2021generating,kumar2020syntax}.
However, little progress has been made in the automatic evaluation of this task. It is even unclear which metric is more reliable among many widely used metrics.  


Most evaluation metrics used in previous paraphrase generation research are not designed for the task itself, but adopted from other evaluation tasks, such as machine translation (MT) and summarization.
However, paraphrase evaluation is inherently different from the evaluation of most other tasks, because a good paraphrase typically obeys two criteria~\cite{gleitman1970phrase,chen2011collecting,bhagat2013paraphrase}: semantic similarity ($Sim$) and lexical divergence ($Div$).
$Sim$ means that the paraphrase maintains similar semantics to the input sentence, whereas $Div$ requires that the paraphrase possesses lexical or syntactic differences from the input.
In contrast, tasks like machine translation have no requirement for $Div$.
It is therefore uncertain whether the metrics borrowed from other tasks perform well in paraphrase evaluation.

In this paper, we revisit automatic metrics for paraphrase evaluation. We collect a list of popular metrics used in recent researches~\cite{kumar2019submodular,feng2021survey,hegde2020unsupervised,sun2021aesop,huang2021generating,kumar2020syntax}, and computed their correlation with human annotation.
Generally, these metrics fall into two categories, i.e., reference-based and reference-free metrics. The former is utilized much more frequently than the latter.


We first empirically quantify the matching degree between metric scores and human annotation, on two datasets of different languages.
Upon both benchmarks, we make comprehensive experiments to validate the reliability of existing metrics. 
Surprisingly, we obtain two important findings: (1) Reference-free metrics better align with human judgments than reference-based metrics on our benchmarks, which is counter-intuitive in related evaluation tasks. (2) Most of these metrics (especially the commonly-used BLEU and Rouge) do not agree well with human evaluation. 

Then we explore the potential reasons behind the above findings through additional experiments.
For the first finding, we demonstrate that the performance comparison between reference-free and reference-based metrics is largely affected by the input-candidate and reference-candidate distance distribution.
Specifically, \emph{reference-free metrics are better because most paraphrase candidates in the testset have larger lexical distances to the reference than to the input, but reference-based metrics may be better for the minority candidates.}
To study the second finding, we design an approach based on attribution analysis~\cite{ajzen1975bayesian,anderson1976bank} to decouple the factors of semantic similarity and lexical divergence.
Our experiments and analysis show that \emph{existing metrics measure semantic similarity well, but tend to ignore lexical divergence}.

Based on our analyses, we propose a new family of metrics named ParaScore for paraphrase evaluation, which takes into account the merits from both reference-based and reference-free metrics and explicitly models lexical divergence.
Extensive experiments show that our proposed metrics significantly outperform the ones employed in previous research.

In summary, our main contributions are:~\footnote{The new dataset and the code of ParaScore is available at the supplementary materials.}
\begin{itemize}[wide=1\parindent,noitemsep, topsep=-1pt]
    \item We observe two interesting findings that disobey conventional wisdom. First, reference-free metrics outperform reference-based ones on our benchmarks. Second, most existing metrics do not align well with human annotation.
    
    \item Underlying reasons behind the above findings are explored through additional experiments and in-depth analysis.
    
    \item Based on the findings and analysis, we propose ParaScore, a family of evaluation metrics for paraphrase generation. They align significantly better with human annotation than existing metrics.
\end{itemize}

\section{Revisiting Paraphrasing Metrics}

\subsection{Settings}
In a standard supervised paraphrase evaluation scenario, we are given an input sentence $X$ and a reference $R$ (the golden paraphrase of $X$). The goal is to evaluate the quality of a paraphrase candidate $C$.

\paragraph{Dataset}
Our experiments selected two benchmarks: Twitter-Para (English) and BQ-Para (Chinese).
Twitter-Para is from the Twitter dataset \cite{xu2014extracting,xu2015semeval}, while BQ-Para is built based on the BQ dataset \cite{chen2018bq}.

Specifically, considering that some metrics may have hyper-parameters, so we use 10\% data in the benchmark as the dev set and tune the hyper-parameters on the dev set. Then the performance of metrics is evaluated on the remaining 90\% data. 
Please refer to Appendix~\ref{tpara} for more details about the two datasets.

\paragraph{Chosen Metrics}
We select the following well known metrics: 
\textbf{BLEU} \cite{papineni2002bleu}, \textbf{ROUGE} \cite{lin2004rouge}, \textbf{METEOR} \cite{banerjee2005meteor}, 
\textbf{BERTScore} \cite{zhang2019bertscore}, and \textbf{BARTScore}~\cite{yuan2021bartscore}. Specifically, we consider two variants of BERTScore: \textbf{BERTScore(B)} and \textbf{BERTScore(R)}, based on BERT~\cite{devlin2019bert} and RoBERTa~\cite{liu2019roberta} respectively. For each metric $M$, we consider its two variants, i.e., a reference-based version and a reference-free version `$M$.Free'.
In the reference-free version, the quality of a candidate $C$ is estimated by $M(C, X)$, where $X$ is the input. Similarly, in the reference-based version, the formula is $M(C, R)$, where $R$ is the reference.

\subsection{Experimental Results}\label{base}

\begin{table}[h]
\centering
\resizebox{\linewidth}{!}{
\begin{tabular}{@{}c|cc|cc@{}}
\toprule[2pt]
\multirow{2}{*}{Metric} & \multicolumn{2}{c}{Twitter-Para} & \multicolumn{2}{|c}{BQ-Para} \\ \cmidrule(l){2-5} 
                        & Pr.           & Spr.          & Pr.           & Spr.          \\ \midrule[2pt]
BLEU-4               &-0.119&-0.104&0.127&0.144              \\
BLEU-4.Free          &\color[HTML]{FE0000}-0.113$\uparrow$&\color[HTML]{FE0000}-0.101$\uparrow$&\color[HTML]{3166FF}0.109$\downarrow$&\color[HTML]{3166FF}0.136$\downarrow$                    \\\midrule
Rouge-1        &0.271    &0.276                  & 0.229         & 0.206         \\
Rouge-1.Free                 &\color[HTML]{FE0000}0.292$\uparrow$ &\color[HTML]{FE0000}0.300$\uparrow$         &\color[HTML]{FE0000}0.264$\uparrow$&\color[HTML]{FE0000}0.232$\uparrow$\\\midrule
Rouge-2        &0.181    &0.144   &0.226 &0.216                   \\
Rouge-2.Free                 &\color[HTML]{FE0000}0.228$\uparrow$
&\color[HTML]{FE0000}0.189$\uparrow$ &\color[HTML]{FE0000}0.252$\uparrow$ &\color[HTML]{FE0000}0.242$\uparrow$
    \\\midrule
Rouge-L        &0.249    &0.239  &0.221 &0.204                   \\
Rouge-L.Free                 &\color[HTML]{FE0000}0.266$\uparrow$ &\color[HTML]{FE0000}0.253$\uparrow$ &\color[HTML]{FE0000}0.260$\uparrow$ & \color[HTML]{FE0000}0.230$\uparrow$  \\\midrule
METEOR         & 0.423        & 0.418    &-&-                     \\
METEOR.Free                  & \color[HTML]{FE0000}0.469$\uparrow$       & \color[HTML]{FE0000}0.471$\uparrow$    &-&-         \\\midrule
BERTScore(B)   &0.470    &0.468         &0.332 &0.322            \\
BERTScore(B).Free            &\color[HTML]{FE0000}0.491$\uparrow$ &\color[HTML]{FE0000}0.488$\uparrow$ &\color[HTML]{FE0000}0.397$\uparrow$&\color[HTML]{FE0000}0.392$\uparrow$
        \\\midrule
BERTScore(R)   &0.368    &0.358    &0.387 &0.376                   \\
BERTScore(R).Free            &\color[HTML]{FE0000}0.373$\uparrow$
& \color[HTML]{FE0000}0.361$\uparrow$ &\color[HTML]{FE0000}0.449$\uparrow$ &\color[HTML]{FE0000}0.438$\uparrow$

\\\midrule
BARTScore      &0.311    &0.306       &0.241&0.230                  \\
BARTScore.Free               &\color[HTML]{3166FF}0.295$\downarrow$
&\color[HTML]{3166FF}0.286$\downarrow$    &\color[HTML]{FE0000}0.282$\uparrow$ &\color[HTML]{FE0000}0.263$\uparrow$

       \\\bottomrule[2pt]
\end{tabular}}
\caption{The Pearson (Pr.) and Spearman (Spr.) correlations between popular metrics and human judgments on two datasets. \textbf{\color[HTML]{FE0000} Red} text (or the text followed by `$\uparrow$') indicates that reference-free metrics are better, whereas \textbf{\color[HTML]{3166FF} blue} text (or the text followed by `$\downarrow$') means the opposite. Please note that we do not apply METEOR to BQ-Para since METEOR is based on the English WordNet \cite{miller1995wordnet}.}
\label{table1}
\end{table}

For each dataset and metric, the score of each sentence in the dataset is calculated by the metric. The obtained scores are then compared with human annotation to check their correlation. The correlation scores, measured by Pearson and Spearman correlations, are reported in Table~\ref{table1}.
Several observations can be made from the table.

\paragraph{Reference-based vs. reference-free}
It can be seen from the table that, for most metrics, their reference-free variants align better with human annotation than their reference-based counterparts.
This indicates that reference-free metrics perform better in the paraphrase generation task, which is somewhat counterintuitive.
More detailed analysis about this observation will be given in Sec~\ref{12}.

\paragraph{Low correlation}
The second observation is that, the correlation between the metrics and human judgments is not high. In other words, most of the commonly-used metrics do not align well with human annotation.
BLEU-4 even shows a negative correlation with human annotation on Twitter-Para.
As the third observation, embedding-based metrics (e.g., BERTScore) tend to outperform ngram-based ones (including the variants of BLEU and Rouge).
The main reason for this lies in the effectiveness of embedding-based metrics in capturing semantic similarity.
Despite the better performance, embedding-based metrics are still far from satisfactory.
On one hand, the results in the table show that the correlation scores for the embedding-based metrics are not high enough.
On the other hand, embedding-based metrics assign a very high score for a candidate if it is the same as the input text. This is an obvious flaw, because it violates the lexical divergence criterion of paraphrasing.
Therefore, we can see obvious drawbacks for both ngram-based and embedding-based metrics.




In summary, we have two findings from the experimental results: (1) Reference-free metrics outperform reference-based ones on our benchmarks. (2) Most of the popular metrics (especially the commonly-used BLEU and Rouge) do not align well with human annotation.

Since the two findings are more or less surprising, some study is necessary to reveal the underlying reasons behind the findings.
We hope the study helps to discover better metrics for paraphrase generation.
In-depth analysis to the two findings are shown in Sec~\ref{12} and Sec~\ref{13} respectively.

\section{Reference-Free vs. Reference-Based}\label{12}

The results in the previous section indicate that reference-free metrics typically have better performance than their reference-based counterpart.
In this section, we investigate this finding by answering the following question:
When and why do reference-free metrics outperform their reference-based variants?

\subsection{The Distance Effect}\label{gap}
Recall that the reference-based and reference-free variants of a metric $M$ calculate the score of a candidate sentence $C$ by $M(C,R)$ and $M(C,X)$ respectively.
Intuitively, as shown in \cite{freitag2020bleu,rei2021references}, if the lexical distance between $C$ and $R$ is large, $M(C, R)$ may not agree well with human annotation.
Therefore, we guess the lexical distance $Dist(C,R)$ between $C$ and $R$ may be an important factor that influences the performance of $M(C,R)$ with respect to human evaluation.

To verify this conjecture, we divide the candidates in a benchmark (e.g., Twitter-Para) into four equal-size groups (group 1 to group 4) according to $Dist(C,R)$,\footnote{Here $Dist$ is measured by normalized edit distance (NED), which is widely used in retrieving translation memory~\cite{cai2021neural,he2021fast}. Its definition is deferred to Appendix~\ref{ned}
.} where elements in group 1 have small $Dist(C,R)$ values.
The performance of several reference-based metrics on such four groups is shown in Figure~\ref{1}\footnote{We can see a counter-intuitive observation that the highest correlation on the subset is lower than the one on the whole set. This is a reasonable statistical phenomenon called Simpson's paradox \cite{wagner1982simpson}.}. It can be seen that when $Dist(C, R)$ grows larger, the performance of the metrics decreases. There is a significant performance drop from group 3 to group 4 when the lexical distance is very large.

\begin{figure}[!h]
    \centering
    \includegraphics[scale=0.038]{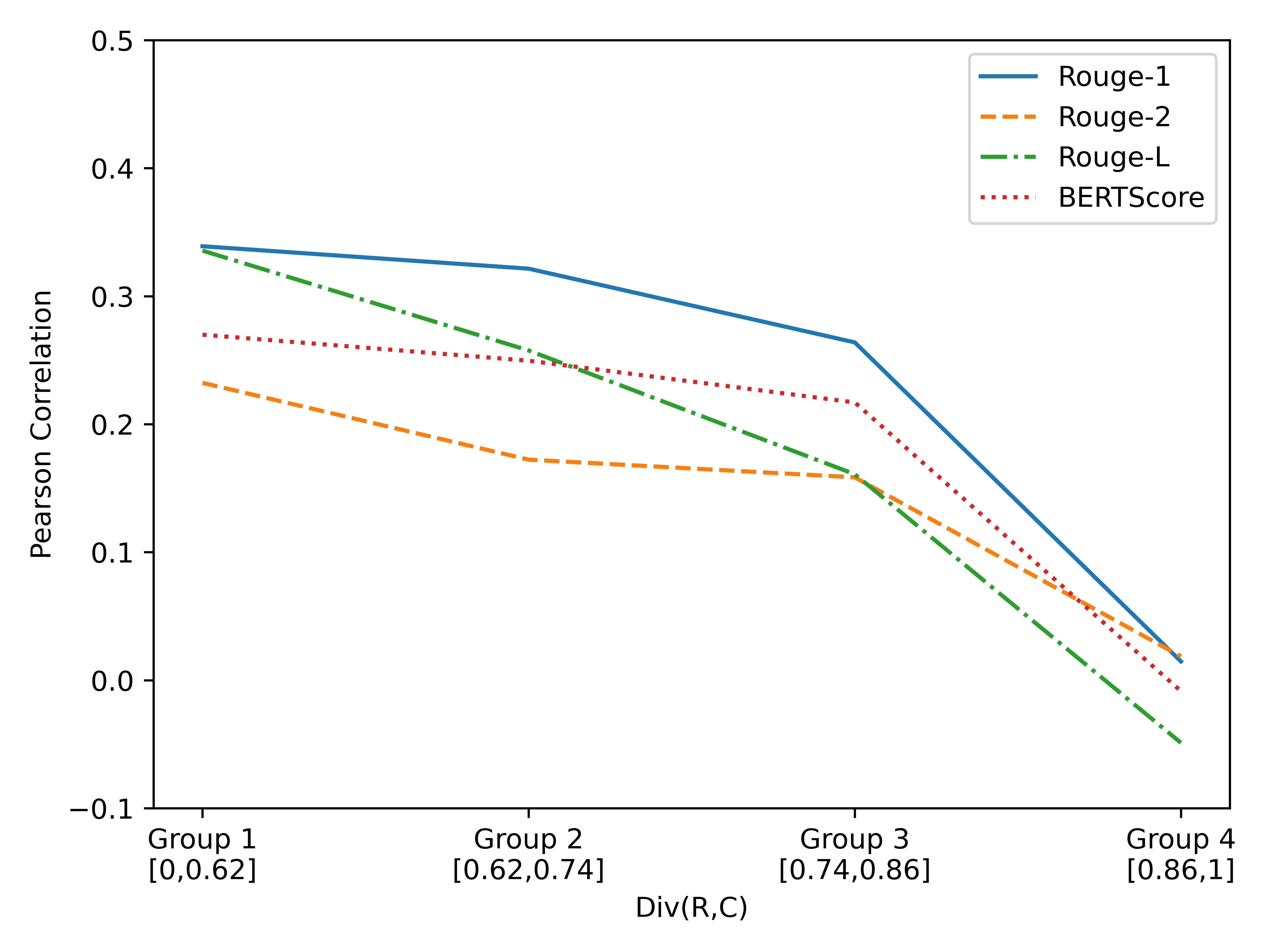}
    \caption{Performance of reference-based metrics significantly degrades as $Dist(R,C)$ becomes large.}
    \label{1}
\end{figure}

\begin{figure}[!h]
    \centering
    \includegraphics[scale=0.038]{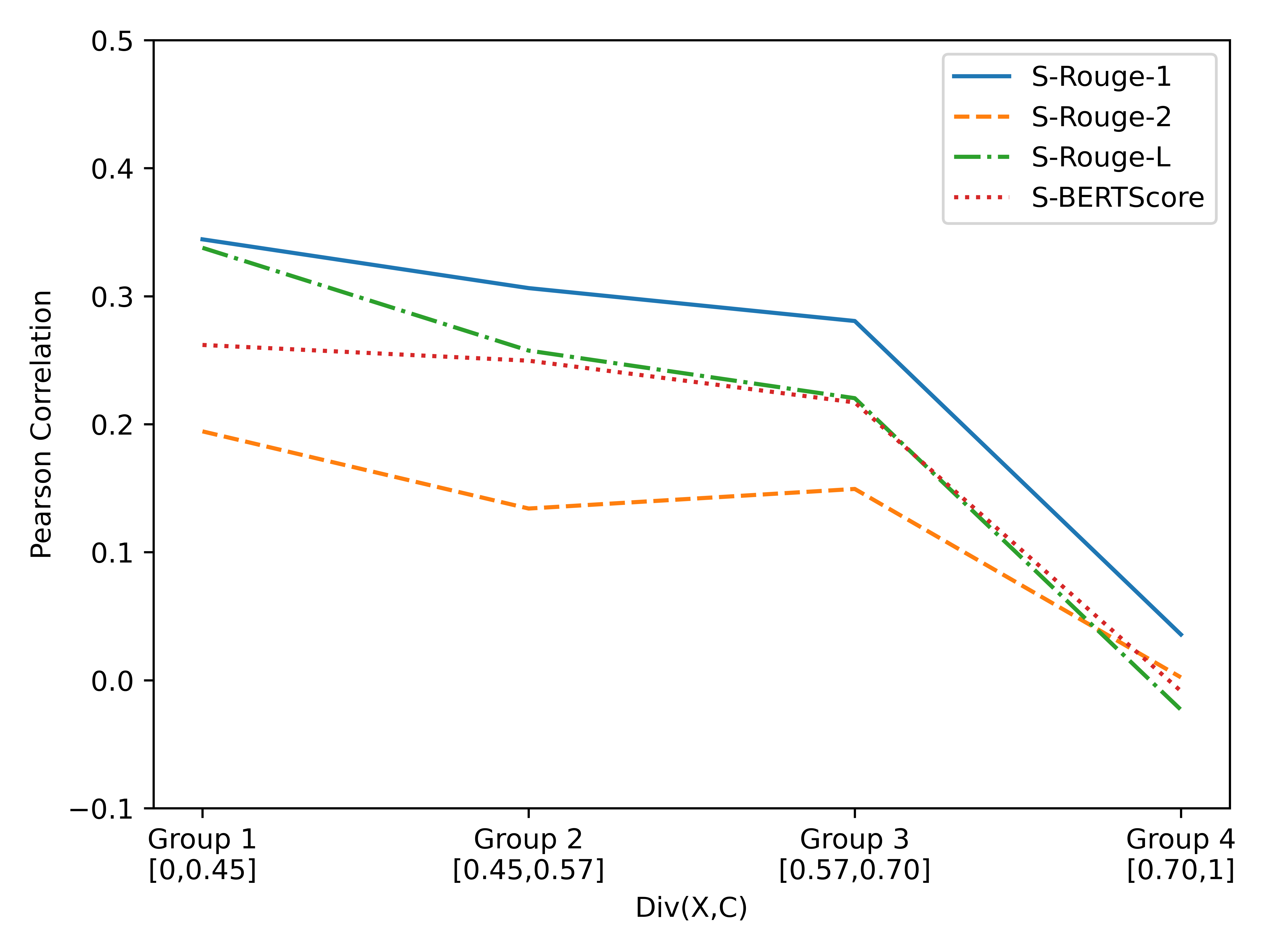}
    \caption{Performance of reference-free metrics significantly degrades as $Dist(X,C)$ becomes large.}
    \label{2}
\end{figure}

Similarly experiments are conducted for reference-free metrics.
We separate the sentences in Twitter-Para into four equal-size groups according to $Dist(C, X)$ and obtain the results in Figure~\ref{2}. Again, the correlation between each metric and human annotation decreases when $Dist(X,C)$ gets larger. A significant performance drop is observed when the lexical distance is very large (see group 4). The above results indicate that small lexical distances are important for both reference-based and reference-free metrics to produce high-quality scores.


\subsection{Average Distance Hypothesis}\label{when}
According to the results in the previous subsection, the average distance from a group of candidates to $R$ or $X$ has a large effect on the performance of a metric on this candidate group.
It is reasonable to further guess that lexical distances also affects the performance comparison between reference-based and reference-free metrics.

Therefore we make the following \textbf{average distance hypothesis}:

\textit{For a group of candidates $G$, a reference-based metric outperforms its reference-free counterpart on $G$ if $Dist(G, R)$ is significantly larger than $Dist(G, X)$}.
\textit{Similarly, the reference-free version is better if $Dist(G, X)$ is greatly larger than $Dist(G, R)$}.

Here $Dist(G, X)$ denotes the average lexical distance from the candidates in $G$ to $X$.


\begin{figure}[!h]
    \centering
    \includegraphics[scale=0.5]{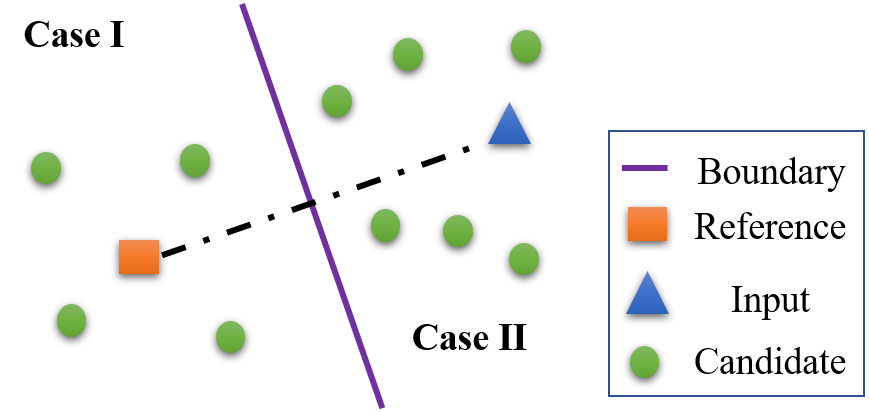}
    \caption{The boundary is the vertical parallel of the `reference-input' line, which separate candidates into two cases. 
    Case {\color[HTML]{6200C9} \uppercase\expandafter{\romannumeral1}} means $Dist(R,C)$ is smaller than $Dist(X,C)$, while Case {\color[HTML]{6200C9} \uppercase\expandafter{\romannumeral2}} means the opposite.}
    \label{333}
\end{figure}

\begin{table}[!h]
\centering
\resizebox{\linewidth}{!}{
\begin{tabular}{@{}c|cccc@{}}
\toprule[2pt]
& \multicolumn{2}{c}{Twitter-Para} & \multicolumn{2}{c}{BQ-Para}\\ \midrule
Metric & {\color[HTML]{6200C9} \uppercase\expandafter{\romannumeral1}} & {\color[HTML]{6200C9} \uppercase\expandafter{\romannumeral2}} & {\color[HTML]{6200C9} \uppercase\expandafter{\romannumeral1}} & {\color[HTML]{6200C9} \uppercase\expandafter{\romannumeral2}}       \\ \midrule[2pt]
RougeL      &0.357   &0.191     &0.352  &0.193\\
RougeL.Free    &0.207   &0.325     &0.319  &0.280\\\midrule
Rouge1      &0.367   &0.223     &0.362  &0.199\\
Rouge1.Free    &0.267   &0.345    &0.308  &0.270\\\midrule
Rouge2      &0.256   &0.120     &0.366  &0.200\\
Rouge2.Free    &0.160   &0.275     &0.283  &0.260\\\midrule
BERTScore   &0.284   &0.162     &0.404  &0.319\\ 
BERTScore.Free &0.191   &0.277     &0.400  &0.417\\\midrule
$\Delta(M.Free, M)$ &-0.110    &+0.132    &-0.044    &+0.079     \\
\bottomrule[2pt]
\end{tabular}}
\caption{The correlation of metrics concerning human annotation on the two parts of Twitter-Para and BQ-Para. $\Delta(M.Free, M)$ denotes the averaged correlation difference between the reference-free metrics ($M.Free$) and the reference-based metrics $M$ per metric. $\Delta(M.Free, M)>0$ indicates the reference-free metric ($M.Free$) is better.}
\label{ccc1}
\end{table}

To validate the above hypothesis, we divide a dataset into two parts (part-I and part-II) according to whether $Dist(C,R)>Dist(X,C)$ or not, as shown in Figure~\ref{333}. 
Then we compare the performance of reference-free and reference-based metrics on the two parts of data.
The performance of reference-free and reference-based metrics on such cases are listed in Table~\ref{ccc1}.
It is clearly shown that reference-based metrics enjoy better performance on part-I, whereas reference-free metrics perform better on part-II.
Such results do verify our average distance hypothesis.

\subsection{Why do Reference-Free Metrics Perform Better on our Benchmarks?}
By employing the average distance hypothesis, we explain why reference-free metrics have higher performance on our datasets.
We calculate the proportion of candidates in {\color[HTML]{6200C9}Case \uppercase\expandafter{\romannumeral1}} and {\color[HTML]{6200C9}Case \uppercase\expandafter{\romannumeral2}} (referring to Figure~\ref{333}) on Para-Twitter and BQ-Twitter.
The results are presented in Table~\ref{cccccc}.
It is shown that there is a larger fraction of Case-II candidates than Case-I on each dataset.
Therefore, according to the average distance hypothesis, it is reasonable to have the observation that reference-free metrics are often better than their reference-based counterparts on both datasets.

\begin{table}[!h]
\centering
\resizebox{\linewidth}{!}{
\begin{tabular}{@{}c|cccc@{}}
\toprule[2pt]
& \multicolumn{2}{c}{Twitter-Para} & \multicolumn{2}{c}{BQ-Para}\\ \midrule
Metric & {\color[HTML]{6200C9} \uppercase\expandafter{\romannumeral1}} & {\color[HTML]{6200C9} \uppercase\expandafter{\romannumeral2}} & {\color[HTML]{6200C9} \uppercase\expandafter{\romannumeral1}} & {\color[HTML]{6200C9} \uppercase\expandafter{\romannumeral2}}       \\ \midrule[2pt]
$\Delta(M.Free, M)$ &-0.110    &+0.132    &-0.044    &+0.079\\
Proportion &46.4\% &53.6\% &15.7\% &84.3\% \\ 
\bottomrule[2pt]
\end{tabular}}
\caption{The proportion of {\color[HTML]{6200C9}Case \uppercase\expandafter{\romannumeral1}} and {\color[HTML]{6200C9}Case \uppercase\expandafter{\romannumeral2}} candidates on Twitter-Para and BQ-Para. A positive $\Delta(M.Free, M)$ means reference-free metrics are better, whereas a negative value indicates that reference-based metrics have better performance.}
\label{cccccc}
\end{table}

\section{Decoupling Semantic Similarity and Lexical Divergence}
In this section, we investigate why most metrics do not align well with human annotation.
\label{13}
\subsection{Attribution Analysis for Disentanglement}\label{11}
As illustrated earlier, a good paraphrase typically obeys two criteria: semantic similarity ($Sim$) and lexical divergence ($Div$).
To seek the reasons behind the low performance of the metrics, we may need to explore how well these metrics perform in terms of each criterion.
However, only one human score is available for a candidate on each dataset. The score is about the overall paraphrasing quality rather than those for a single criterion (either semantic similarity or lexical divergence). 

In this section, we propose an approach to decouple the performance of the metrics in terms of each criterion. 
This proposed approach is inspired by attribution analysis~\cite{anderson1976bank,ajzen1975bayesian} and its key idea is to analyze the attribution of one component (or dimension) while controlling the attributions from other components (or dimensions).

Applying attribution analysis to our scenario, we construct a subset $\mathcal{S} = \{(X, C_j, C_k)\}$, where $(C_j, C_k)$ is a paraphrase candidate pair for an input sentence $X$, such that the difference between $C_j$ and $C_k$ on one criterion ($Sim$ or $Div$) is significant but the difference on the other criterion is close to zero. As a result, on such a subset $\mathcal{S}$, the difference of human score between $C_j$ and $C_k$ is mainly attributed by the interested criterion. Then we can measure the correlation between human scores and a metric in the specific criterion.

\begin{table}[h]\small
\centering
\begin{tabular}{@{}c|cc|cc@{}}
\toprule[2pt]
\multirow{2}{*}{} & \multicolumn{2}{c}{Twitter-Para} & \multicolumn{2}{|c}{BQ-Para} \\ \cmidrule(l){2-5} 
& $\mathcal{S}_\text{sim}$ & Base          & $\mathcal{S}_\text{sim}$ & Base  \\ \midrule[2pt]
\#num  &583 &9158  &200 &5156  \\
$\rho$               &0.805  &0.345   &0.629  &0.394  \\
\bottomrule[2pt]
\end{tabular}
\caption{Pearson correlation of $\Delta S$ and $\Delta h$ on $\mathcal{S}_\text{sim}$ compared with that on paraphrase pairs filtered by only Eq.(1) only (Base). The results also demonstrate the necessity of the constraint Eq.(2).}
\label{abcd}
\end{table}

Since there are no ground truth measures for $Sim$ and $Div$, we use normalized edit distance (NED) and SimCSE~\cite{gao2021simcse} as the surrogate ground truth of $Div$ and $Sim$ respectively.
They are chosen for two reasons.
First, they are widely used and proven to be good for measuring $Div$ and $Sim$,. Second, they are not used as the metrics for paraphrase evaluation in this paper. Therefore, the potential unfairness is reduced.\footnote{For example, if we use BERTScore to compute $Sim$, the statistics on $\mathcal{S}$ may be biased to BERTScore and thus becomes unfair for other metrics.}

\subsection{Performance in Capturing $Sim$}\label{sim}

Formally, suppose the subset $\mathcal{S}_\text{sim}$ denotes all $(X, C_j, C_k)$ satisfying the following constraints:
\begin{equation}
\begin{aligned}\label{1}
       |Dist(X, C_j)-Dist(X,C_k)| &\leq \eta_{1}  \\
       |Sim(X, C_j)-Sim(X,C_k)| &\geq \eta_{2}  
\end{aligned}
\end{equation}
\noindent where $Dist$ is a distance function for calculating $Div$, $\eta_{1}$ is set as 0.05 and $\eta_{2}$ is 0.15. \footnote{Intuitively, the disentanglement effect would be better if $\eta_1$ is more close to zero and $\eta_2$ is much larger. 
 However, this leads to the limited size of $\mathcal{S}_\text{sim}$ due to the contradictory between $Sim$ and $Div$, and hence the statistical correlation on $\mathcal{S}_\text{sim}$ is not significant.}

In addition, we define two quantities for each tuple $(X, C_j, C_k)$ from $\mathcal{S}_\text{sim}$ as follows:
\begin{align}\label{eq1}
&\Delta S=Sim(X,C_{j})-Sim(X,C_{k}) \\ &\Delta h=h(X,C_{j})-h(X,C_{k}) 
\end{align}
where $h()$ refers to the human score.
Then we measure the correlation between $\Delta S$ and $\Delta h$ on $\mathcal{S}_\text{sim}$, and the results are shown in Table~\ref{abcd}. 
It can be seen that the correlation is much higher on $\mathcal{S}_\text{sim}$ compared with that on all paraphrase pairs, indicating good disentanglement on $\mathcal{S}_\text{sim}$.
As $\mathcal{S}_\text{sim}$ is proper to demonstrate how well a metric captures semantic similarity, we call it {\bf semantic-promoted data}. 




\begin{table}[t]\small
\centering
\begin{tabular}{@{}c|cc@{}}
\toprule[2pt]
{Metric} & Twitter-Para           & BQ-Para                    \\ \midrule[2pt]
BLEU-4.Free  &0.067&0.372                     \\
Rouge-1.Free &0.574&0.430                    \\
Rouge-2.Free  &0.400&0.350                   \\
Rouge-L.Free  &0.481&0.388                   \\
METEOR.Free  &0.499&-                   \\
BERTScore(B).Free &0.785&0.576                 \\
BARTScore.Free  &0.797&0.552                  \\
Sim  &0.805&0.629                  \\\bottomrule[2pt]
\end{tabular}
\caption{Pearson correlation of $\Delta M$ and $\Delta h$ on $S_\text{sim}$, the `semantic-promoted data'. This is an example to show that paraphrase quality does not increase as lexical divergenceincreases.}
\label{similarity}
\end{table}

To investigate how well existing metrics capture semantic similarity, we add an extra definition:
\begin{equation}
    \Delta M = M(X,C_{j})-M(X,C_{k})
\end{equation}
where $M$ is a reference-free metric. Then we measure the correlation between $\Delta M$ and $\Delta h$ on the semantic-promoted data, and get the results in Table~\ref{similarity}.
The results suggest that the embedding-based metrics (i.e., BERTScore.Free) significantly outperform word-overlap metrics (i.e., BLEU.Free) in capturing semantic similarity.
Overall, the results show that some metrics perform pretty well in capturing semantic similarity.

\subsection{Performance in Capturing $Div$}\label{4.3}

Similarly, to analyze the ability of metrics in capturing $Div$, we exchange $Dist$ with $Sim$ in Eq~\ref{1} and obtain a subset of tuples named $\mathcal{S}_\text{div}$ ($\eta_{1}=0.05$ and $\eta_{2}=0.10$). In this case, the principal attribution on $\mathcal{S}_\text{div}$ is lexical divergence. 
In addition, we define $\Delta D$ as follows:
\begin{align}\label{eq22}
&\Delta D=Dist(X,C_{j})-Dist(X,C_{k}) 
\end{align}
Then we conduct analyses on $\mathcal{S}_\text{div}$ to examine the effect of disentanglement for lexical divergence.
It is interesting that the correlation between $\Delta D$ and $\Delta h$ on $\mathcal{S}_\text{div}$ is almost zero, which indicates that higher distance scores does not guarantee better paraphrasing. 
This fact is in line with previous findings \cite{bhagat2013paraphrase}.
Let's explain by the examples in Table~\ref{div}. It is reasonable for candidate $C_1$ to get a low human annotation score due to its small lexical distance to the input $X$.
Though $C_3$ has a larger distance to $X$ than $C_2$, they are assigned the same annotation score, possibly because both $C_2$ and $C_3$ are good enough in terms of $Div$ from the viewpoint of human annotators.
Such results show that when the distance is large (i.e., beyond a threshold), $Div$ does not correlate well with human score $h$.
\begin{table}[h]\small
\centering
\begin{tabular}{@{}c|l|ccc@{}}
\toprule[2pt]
Type & \multicolumn{1}{c|}{Text}                                                     & $Dist$   & $h$   \\ \midrule[2pt]
$X$    & \begin{tabular}[c]{@{}l@{}}NLP is a potential research field\end{tabular} & -       & -   \\ \midrule
$C_1$    & \begin{tabular}[c]{@{}l@{}}NLP is a promising research field\end{tabular} & 0.21  & 0.4 \\
$C_2$    & \begin{tabular}[c]{@{}l@{}}NLP is a promising  study area\end{tabular}     & 0.53  & 1.0 \\ 
$C_3$    & \begin{tabular}[c]{@{}l@{}}The NLP field has high potential\end{tabular}  & 0.79  & 1.0 \\
 \bottomrule[2pt]
\end{tabular}
\caption{$X$ and $C$ refer to the input and candidate. This example shows that paraphrase quality annotated by human ($h$) does not always increase as the lexical divergence ($Dist$) increases.}
\label{div}
\end{table}

We modify our decoupling strategy by further dividing $\mathcal{S}_\text{div}$ into two parts according to a distance threshold. We define $d$ as follows:
\begin{equation}
    d(j,k) = \min(Dist(X,C_{j}),Dist(X,C_{k}))
\end{equation}
where $d(j,k)$ represents the minimum $Dist$ score in $(X, C_j, C_k)$. We use 0.35 as the threshold to split $\mathcal{S}_\text{div}$, with $\mathcal{S}_\text{div1}$ containing all the tuples satisfying $d(j,k)<=0.35$, and $\mathcal{S}_\text{div2}$ containing other tuples.
The Pearson correlation of $\Delta D$ and $\Delta h$ on the two subsets are listed in Table~\ref{abc}.
According to the results, the correlation is high on $\mathcal{S}_\text{div1}$ but almost zero on $\mathcal{S}_\text{div2}$.
This is consistent with our intuition that candidates with larger $Div$ scores tend to have higher quality when the distances are under a threshold. However, increasing $Div$ scores does not improve quality when the distances exceed a threshold.

\begin{table}[h]\small
\centering
\begin{tabular}{@{}c|cc|cc@{}}
\toprule[2pt]
\multirow{2}{*}{} & \multicolumn{2}{c}{Twitter-Para} & \multicolumn{2}{|c}{BQ-Para} \\ \cmidrule(l){2-5} 
& $\mathcal{S}_\text{div1}$ & $\mathcal{S}_\text{div2}$          & $\mathcal{S}_\text{div1}$ & $\mathcal{S}_\text{div2}$       \\ \midrule[2pt]
\#num &192 &3876 &290 &6217  \\
$\rho$             &0.635  &0.021    &0.655  &0.025  \\
\bottomrule[2pt]
\end{tabular}
\caption{Pearson correlation of $\Delta D$ and $\Delta h$ on two partitions of $\mathcal{S}_\text{div}$ controlled by a threshold ($0.35$).}
\label{abc}
\end{table}

The correlation between $\Delta M$ and $\Delta h$ on $\mathcal{S}_\text{div1}$ are shown in Table~\ref{diversity}.
It is shown that the correlation scores for all the metrics (except for the Dist function itself) are negative, which means the metrics tend to have opposite judgments with human annotators about the paraphrasing quality for the candidates in $\mathcal{S}_\text{div1}$.
\begin{table}[!h]\small
\centering
\begin{tabular}{@{}c|cc@{}}
\toprule[2pt]
\multirow{1}{*}{Metric}   & Twitter-Para           & BQ-Para                    \\ \midrule[2pt]
BLEU-4.Free  &-0.197&-0.075                     \\
Rouge-1.Free &-0.385&-0.334                    \\
Rouge-2.Free  &-0.377&-0.308                   \\
Rouge-L.Free  &-0.426&-0.514                   \\
METEOR.Free  &-0.233&-                   \\
BERTScore(B).Free &-0.424&-0.347                 \\
BARTScore.Free  &-0.187&-0.263                  \\
NED  &0.635&0.655                  \\\bottomrule[2pt]
\end{tabular}
\caption{Pearson correlation of $\Delta M$ and $\Delta h$ on $\mathcal{S}_\text{div1}$.}
\label{diversity}
\end{table}


\section{New Metric: ParaScore}
\subsection{ParaScore}
Inspired by previous experiments and analyses, we propose a new metric named ParaScore, as below,
\begin{multline}
 \label{eq:parascore}
  \text{ParaScore}= \max({Sim}(X,C),{Sim}(R,C))+ \\
  \omega\cdot {DS}(X,C)  
\end{multline}
\noindent where $\omega$ is a hyper-parameter in our experiments, $\max({Sim}(X,C),{Sim}(R,C))$ is motivated by the analysis in \S \ref{when}, and $DS$ is defined as a sectional function to model lexical divergence (referring to the analysis in \S \ref{4.3}):
\begin{equation}\label{sec}
{DS}(X,C)=\left\{
\begin{array}{cl}
\gamma &  \text{d} \textgreater{} \gamma \\
\text{d}\cdot\frac{\gamma+1}{\gamma}-1  & 0 \leq \text{d} \leq \gamma \\
\end{array} \right.
\end{equation}
\noindent where $\gamma$ is a hyper-parameter, \text{d}=$Dist(X, C)$, which can be any proper distance metric. In our experiments, $Sim$ and $Div$ are respectively instantiated by BERTScore and NED~\footnote{Note that there may be other advanced metrics to instantiate $Sim$ (e.g., SimCSE) and other heuristic combination (e.g., weighted geometric mean) methods, which we leave as future work.}, and $\gamma$ is fixed as 0.35.

ParaScore defined in Eq.~\eqref{eq:parascore} involves the reference $R$ and thus it is a reference-based metric. It is natural to extend ParaScore to its reference-free version {\bf ParaScore.Free} by removing $R$ as follows:
\begin{equation*}
    \text{ParaScore.Free}= {Sim}(X,C)+\omega\cdot {DS}(X,C). 
\end{equation*}




\begin{table}[t]
\resizebox{\linewidth}{!}{
\begin{tabular}{@{}c|cc|cc@{}}
\toprule[2pt]
\multirow{2}{*}{Metric} & \multicolumn{2}{c}{Twitter-Para} & \multicolumn{2}{c}{BQ-Para} \\ \cmidrule(l){2-5} 
                        & Pearson        & Spearman   & Pearson        & Spearman           \\ \midrule[2pt]
BERTScore(B)            & 0.470         & 0.468         &0.332          &0.322          \\
BERTScore(R)            & 0.368         & 0.358         & 0.387         & 0.376          \\
BARTScore               & 0.311         & 0.306         &0.260          &0.246          \\ 
iBLEU(0.2) &0.013 &0.033 &0.155 &0.139\\ 
BERTScore(B).Free        & 0.491   & 0.488  &0.397          &0.392            \\
BERT-iBLEU(B,4) &0.488 &0.485 &0.393&0.383\\ 
ParaScore &\bf 0.522 &\bf0.523   &\bf0.492 &\bf0.489\\
ParaScore.Free &0.492 &0.489   &0.398 &0.393\\
\bottomrule[2pt]
\end{tabular}}
\resizebox{\linewidth}{!}{
\begin{tabular}{@{}c|cc|cc@{}}
\toprule[2pt]
\multirow{2}{*}{Metric} & \multicolumn{2}{c}{Twitter(Extend)} & \multicolumn{2}{c}{BQ-Para(Extend)} \\ \cmidrule(l){2-5} 
                        & Pearson        & Spearman   & Pearson        & Spearman           \\ \midrule[2pt]
BERTScore(B)            & 0.427         & 0.432         &0.248          &0.267          \\
BERTScore(R)            & 0.334         & 0.329         & 0.299         & 0.317          \\
BARTScore               & 0.280         & 0.276         &0.199          &0.206          \\ 
iBLEU(0.2) &0.011 &0.032 &0.129 &0.121\\ 
BERTScore(B).Free        & 0.316   & 0.419  &0.230          &0.312            \\
BERT-iBLEU(B,4) &0.327 &0.416 &0.221&0.303\\ 
ParaScore &\bf0.527 &\bf0.530   &\bf0.510&\bf0.442 \\
ParaScore.Free &0.496&0.495 &0.487 &0.428 \\
\bottomrule[2pt]
\end{tabular}}
\caption{The Pearson (Pr.) and Spearman (Spr.) correlations on two benchmarks. Specifically, we highlight the best performance with \textbf{Bold numbers}.  BERT-iBLEU(B,4) means the encoder is BERT and $\beta$ is 4. iBLEU(0.2) indicates $\alpha$ is set as 0.2. }
\label{tablebbb}
\end{table}

\subsection{Experimental Results}
\paragraph{Benchmarks and baselines}
Experiments are conducted on four datasets: Twitter-Para, BQ-Para, and the extended version of them.
The extended version of each dataset is built by adding 20\% of the input sentences as candidates.
They are called \textbf{Twitter(Extend)} and \textbf{BQ(Extend)} respectively.
Since the newly added candidates are input sentences, according to the requirements of paraphrasing, their annotation scores are 0.
The goal of adding the extended version of the datasets is to test the robustness of different metrics on various data distributions.
In addition to the baselines in previous sections, we add two more baselines: \textbf{BERT-iBLEU} \cite{niu2021unsupervised} and \textbf{iBLEU} \cite{siddique2020unsupervised,liu2020unsupervised}, whose details are listed in Appendix~\ref{ibleu}. 
\paragraph{Performance comparison}
The performance of each metric on the four datasets are listed in Table~\ref{tablebbb}.
Several observations can be made.
First of all, ParaScore performs significantly better than all the other metrics on all the datasets.
It is also shown that ParaScore is much more robust than other metrics.
Second, on both Twitter-Para and BQ-Para, BERT-iBLEU performs worse than vanilla BERTScore.
Note that BERT-iBLEU \cite{niu2021unsupervised} also considers lexical divergence, and it applies a harmonic weight mean of BERTScore (for semantic similarity) and -BLEU.Free (for lexical divergence). However, according to results in Table~\ref{tablebbb}, it is only comparable to BERTScore.Free or even worse. This further indicates that 1) the weighted harmonic mean formation is sub-optimal, 2) the sectional threshold is important as discussed in \S 4.3, making the performance comparable to BERTScore.Free in most cases, as shown in Appendix~\ref{bertibleu}.

\begin{table}[!h]
\resizebox{\linewidth}{!}{
\begin{tabular}{@{}c|cc|cc@{}}
\toprule[2pt]
\multirow{2}{*}{Metric} & \multicolumn{2}{c}{Twitter(Extend)} & \multicolumn{2}{c}{BQ-Para(Extend)} \\ \cmidrule(l){2-5} 
                        & Pr.           & Spr.          & Pr.           & Spr.          \\ \midrule[2pt]
ParaScore & \bf 0.527 & \bf0.530 & \bf 0.510 & \bf0.442
\\
ParaScore w/o thresh            & 0.358         & 0.450&0.266&0.333                  \\
ParaScore w/o max            &0.496         &0.495 &0.487 &0.428                  \\
ParaScore w/o DS            & 0.349         & 0.450 & 0.249 &0.326  \\ \bottomrule[2pt]
\end{tabular}}
\caption{Ablation study on the ParaScore. ParaScore w/o thresh means removing the sectional formation defined in Eq~\ref{sec}. ParaScore w/o DS means removing the lexical divergencescore.}
\label{tablebbbb}
\end{table}

\paragraph{Ablation study}
We study of effect of three factors of ParaScore: the max function, the DS function for divergence, and the threshold mechanism in Equ~\eqref{sec}.
The results are listed in Table~\ref{tablebbbb}.
By comparing ParaScore with `ParaScore w/o DS', we can see that ParaScore significantly degrades when removing $DS$ or its sectional version, which confirms the effectiveness of $DS$ and the sectional function for $DS$. 
These findings demonstrate that a sectional function for $Div$ is beneficial for paraphrase evaluation.
According to the results, all of the above listed factors are essential for the effectiveness of ParaScore.

\paragraph{Discussion}
According to Table~\ref{diversity}, we can observe that existing metrics do not well consider the lexical divergence, including BERTScore.Free.
On the two original benchmarks, as shown in Table~\ref{tablebbb}, BERTScore.Free is still competitive with ParaScore.Free, which explicitly models lexical divergence. This fact seems to disagree with the human evaluation guideline that lexical divergence is also important. Therefore, these results may reveal a potential drawback in the original benchmarks: They overlook the role of lexical divergence. Although the extended version of both benchmarks alleviates such a drawback to some extent, it introduces divergence into both datasets in a toy manner by copying the inputs rather than in a natural manner. It would be important to build a better benchmark for paraphrase evaluation in the future.


\section{Related Work}

Most previous works conduct paraphrase evaluation by the reference-based MT metrics from the popular tasks similar to paraphrase generation such as machine translation \cite{bannard2005paraphrasing,callison2008syntactic,cohn2008constructing,kumar2020syntax,goyal2020neural,sun2021aesop,huang2021generating}. However, paraphrase evaluation is different from these tasks: the paraphrase should possess lexical or syntactic differences toward the input sentence, which is not emphasized in these tasks. 

Generally, the metrics in paraphrase evaluation can be divided into two kinds: reference-free and reference-based metric. Most reference-based metrics include BLEU \cite{papineni2002bleu}, Rouge \cite{lin2004rouge}, and METEOR \cite{banerjee2005meteor}. In addition, the reference-free of these metrics have also been used: Self-BLEU \cite{shu2019generating} measures the BLEU score between the generated paraphrase and input sentence. Moreover, the iBLEU \cite{choshen2018automatic} score penalizes repeating the input sentence in the generated paraphrase. BERT-iBLEU \cite{zhou2021paraphrase} takes the weighted harmonic mean of the BERTscore \cite{zhang2019bertscore} and one minus self-BLEU.
Previous works commonly utilize reference-based metrics in evaluation, in this paper, we also pay attention to the overlooked reference-free metrics. 

The difference between the existing works and our work is obvious. 
Existing works mainly employ these metrics to evaluate the paraphrases generated from a model. However, the reliability of existing paraphrase metrics has not been evaluated comprehensively. Thus, we prepare two paraphrase evaluation benchmarks (Chinese and English) and conduct comprehensive experiments to compare existing metrics' performance on these benchmarks. In particular, based on the empirical findings, this paper proposes a new framework for paraphrase evaluation.


\section{Conclusion}
This paper first reviews the reliability of existing metrics for paraphrasing evaluation by investigating how well they correlate with human judgment. Then, we find two interesting findings and further ask two questions behind them that are overlooked by the community: (1) why do reference-free metrics outperform reference-based ones? (2) what is the limitation of existing metrics? We deliver detailed analyses of such two questions and present the explanation by disentangling paraphrase quality. Based on our analyses, finally, we propose ParaScore (with both reference-based and reference-free implementations) for paraphrase evaluation, and its effectiveness is validated through comprehensive experiments. In addition, we call for building better benchmarks which can faithfully reflect the importance of lexical divergence in paraphrase evaluation; we hope it will shed light on the future direction.

\section*{Limitation}
One limitation in this paper is that it does not provide a perfect benchmark which remarkably reflects the importance of lexical divergence in a natural way rather than the heuristic way used in the experiments. Creating such a benchmark would be important for future studies on paraphrase evaluation. It is also interesting to examine the potential benefits of the proposed ParaScore on such a benchmark. 

\section*{Ethical Considerations}
The datasets used in this paper will not pose ethical problems. 
For the Twitter-Para dataset, it is a publicly available dataset. For the BQ-Para dataset, its inputs are from the public dataset BQ and we recruited five annotators to manually annotate the quality of paraphrases with the proper pay.

\bibliography{acl_latex}
\bibliographystyle{acl_natbib}

\appendix

\section{Details of Twitter-Para}\label{tpara}
Our Twitter-Para is a pre-processed dataset based on \cite{xu2014extracting,xu2015semeval}. In the original dataset \cite{xu2014extracting,xu2015semeval}, there are some input sentences that have no corresponding references, so we drop such input-candidate pairs to create Twitter-Para. Specifically, the human-annotated score ranges from 0$\sim$1.0, where higher scores mean better quality. The basic statistics of Twitter-Para are listed in Table~\ref{para}. 

\begin{table}[h]
\resizebox{\linewidth}{!}{
\begin{tabular}{@{}cccc@{}}
\toprule
\#input & \#candidate & \#reference & avg candidate \\ \midrule
761      & 7159        & 761         & 9.41          \\ \bottomrule
\end{tabular}}
\caption{The statistics of Twitter-Para. There are 761 input sentences and each input sentence corresponds to one standard reference. Besides, there are 7159 paraphrase candidates totally, and each input sentence owns 9.41 paraphrase candidates averagely.}
\label{para}
\end{table}


\section{Details of BQ-Para}\label{bqpara}
Considering the absence of Chinese paraphrase evaluation benchmarks, we build BQ-Para based on the BQ dataset. We select 550 sentences as input sentences from BQ-dataset. Each sentence owns a manually-written reference and also owns ten candidates. Specifically, such candidates are generated by popular paraphrase generation algorithms. Then, for such a candidate, given the input sentence, we hire professional annotators to provide a score between $0-1.0$ to reflect its paraphrase quality. The basic statistics of BQ-Para are listed in Table~\ref{bpara}. 

\begin{table}[!h]
\resizebox{\linewidth}{!}{
\begin{tabular}{@{}cccc@{}}
\toprule
\#input & \#candidate & \#reference & avg candidate \\ \midrule
550      & 5550        & 550         & 10          \\ \bottomrule
\end{tabular}}
\caption{The statistics of BQ-Para. There are 550 input sentences and each input sentence corresponds to one standard reference. Besides, there are 5550 paraphrase candidates totally, and each input sentence owns 10 paraphrase candidates averagely.}
\label{bpara}
\end{table}


\section{Definition of normalized edit distance}\label{ned}
Given two sentences $\mathbf{x}$ and $\mathbf{x}^{i}$, the definition of normalized edit score is defined as follows:
\begin{equation}
NED = \frac{\operatorname{dist}\left(\mathbf{x}, \mathbf{x}^{i}\right)}{\max \left(|\mathbf{x}|,\left|\mathbf{x}^{i}\right|\right)}
\end{equation}
where $|\mathbf{x}|$ is the length of sentence $\mathbf{x}$.

\section{Definition of BERT-iBLEU and iBLEU}\label{ibleu}
BERT-iBLEU is defined as follows:
\begin{equation}\scriptsize
\begin{aligned}
\text { BERT-iBLEU } &=\frac{\beta+1.0}{\beta\cdot\text{BERTScore}^{-1}+1.0\cdot(1-\text{SelfBLEU}) ^{-1}} \\
\text {SelfBLEU}&=\mathrm{BLEU}(\text{input},\text{candidate})
\end{aligned}
\end{equation}
where $\beta$ is a constant (usually set as 4).

iBLEU is a hybrid metric that computes the difference between BLEU and SelfBLEU, which is defined as follows:
\begin{equation}
    \text{iBLEU}=\text{BLEU}-\alpha\cdot\text{SelfBLEU}
\end{equation}
where $\alpha$ is a constant (usually set as 0.3).

\section{A detailed analysis towards BERT-iBLEU}\label{bertibleu}
Principally, we can formulate any existing metrics into the combination of semantic similarity (Sim) and lexical divergence(Div), including BERT-iBLEU. Firstly, we recall the definition of BERT-iBLEU:
\begin{equation}\scriptsize
\begin{aligned}
\text { BERT-iBLEU } &=\frac{\beta+1.0}{\beta\cdot\text{BERTScore}^{-1}+1.0\cdot(1-\text{SelfBLEU}) ^{-1}} \nonumber
\end{aligned}
\end{equation}
Naturally, we re-write BERT-iBLEU as the following formation:
\begin{equation}
\begin{aligned}
\text { BERT-iBLEU } &=\frac{\beta+1.0}{\beta\cdot\text{Sim}^{-1}+\cdot(\text{Div}) ^{-1}} \nonumber
\end{aligned}
\end{equation}
where Sim represents the BERTScore and Div denotes (1-SelfBLEU). Though such a formation indeed contains both lexical divergence and semantic similarity, it can not gaurantee that BERT-iBLEU is a good paraphrase metric that serves as a human-like automatic metric. Existing work \cite{niu2021unsupervised} only shows that it outperforms n-gram-based metrics. The following experiments demonstrate an interesting conclusion: \emph{BERT-iBLEU consistently performs worse than SelfBERTScore}, and then we present our analysis. The results are demonstrated in Table~\ref{worse}, from where we can see that BERT-iBLEU(B) consistently under-perform than BERTScore(B).

\begin{table}[h]
\resizebox{\linewidth}{!}{
\begin{tabular}{@{}c|cc|cc@{}}
\toprule[2pt]
\multirow{2}{*}{Metric} & \multicolumn{2}{c}{Twitter-Para} & \multicolumn{2}{c}{BQ-Para} \\ \cmidrule(l){2-5} 
                        & Pr.           & Spr.          & Pr.           & Spr.          \\ \midrule
BERTScore(B).Free   &0.491&0.488&0.397&0.392     \\
BERT-iBLEU(B,4) &0.488 &0.485 &0.393 &0.383\\
BERT-iBLEU(B,5) &0.490 &0.488 &0.395 &0.392\\
BERT-iBLEU(B,10) &0.490 &0.488 &0.396 &0.389\\
\bottomrule[2pt]
\end{tabular}}
\caption{The Pearson (Pr.) and Spearman (Spr.) correlations of vanilla BERTScore and BERT-iBLEU. We can see BERT-iBLEU consistently under-perform vanilla BERTScore on both benchmarks.}
\label{worse}
\end{table}

To explain such interesting results, we re-write BERT-iBLEU as follows:
\begin{equation}
\begin{aligned}
\text {BERT-iBLEU} &=\frac{\beta+1.0}{\beta\cdot\text{Sim}^{-1}+\cdot(\text{Div}) ^{-1}} \\
&=\frac{\beta \cdot \text{Sim} \cdot \text{Div} + \text{Sim} \cdot \text{Div}}{\beta \cdot \text{Div} + \text{Sim}}\\
&=\text{Sim}+\frac{\text{Sim} \cdot \text{Div} - \text{Sim}^{2}}{\beta \cdot \text{Div} + \text{Sim}} \nonumber
\end{aligned}
\end{equation}
As we can see, BERT-iBLEU can be decoupled into two terms Sim and $\frac{\text{Sim} \cdot \text{Div} - \text{Sim}^{2}}{\beta \cdot \text{Div} + \text{Sim}}$ (We denote it as term `Mix'). According to the analysis in our paper, after removing the Sim, the remaining part, the `Mix' term should be able to reflect diversity. However, the `Mix' term does not represent meaningful aspects of paraphrase quality. Specifically, we investigate the correlation between the `Mix' term and human annotation, only resulting in correlations close to zero, indicating that the `Mix' term is improper since there is nearly no correlation between it and human annotation. Overall, BERT-iBLEU owns an improper combination of semantic similarity and diversity.

\end{document}